\documentclass[10pt,twocolumn,letterpaper]{article}

\usepackage[pagenumbers]{cvpr} 

\usepackage{graphicx}
\usepackage{mmstyles}
\usepackage{amsmath}
\usepackage{amssymb}
\usepackage{booktabs}
\usepackage{multirow}

\usepackage{amsmath,amsfonts,bm}

\def\eqref#1{equation~(\ref{#1})}

\def\1{\bm{1}}

\def\vf{{\bm{f}}}

\def\vx{{\bm{x}}}

\def\mF{{\bm{F}}}

\def\mI{{\bm{I}}}

\def\mX{{\bm{X}}}

\DeclareMathAlphabet{\mathsfit}{\encodingdefault}{\sfdefault}{m}{sl}
\SetMathAlphabet{\mathsfit}{bold}{\encodingdefault}{\sfdefault}{bx}{n}

\def\gN{{\mathcal{N}}}

\usepackage[pagebackref,breaklinks,colorlinks]{hyperref}

\usepackage[capitalize]{cleveref}
\crefname{section}{Sec.}{Secs.}
\Crefname{section}{Section}{Sections}
\Crefname{table}{Table}{Tables}
\crefname{table}{Tab.}{Tabs.}

\def\cvprPaperID{6089} 
\def\confName{CVPR}
\def\confYear{2023}

\begin{document}

\title{Controllable Mesh Generation Through Sparse Latent Point Diffusion Models}

\newcommand{\AuthorSpace}{\hspace{1.2em}}
\author{%
Zhaoyang Lyu$^{1}$\thanks{Equal Contribution.} \AuthorSpace{} Jinyi Wang$^{1,2*}$\AuthorSpace{} Yuwei An$^{1,4}$\AuthorSpace{} Ya Zhang$^{1,2}$\AuthorSpace{} Dahua Lin$^{1,3}$\AuthorSpace{} Bo Dai$^{1}$ \\
$^1$Shanghai AI Laboratory \hspace{1em}
$^2$Shanghai Jiao Tong University\\
$^3$The Chinese University of Hong Kong \hspace{1em}
$^4$Tsinghua University \\
\texttt{lyuzhaoyang@link.cuhk.edu.hk, jinyi.wang@sjtu.edu.cn} \\
\texttt{anyuwei@pjlab.org.cn, ya\_zhang@sjtu.edu.cn}\\
\texttt{dhlin@ie.cuhk.edu.hk, daibo@pjlab.org.cn}
}
\maketitle
\begin{figure*}[h!]
\vspace{-3em}
\centering
\includegraphics[width=1\textwidth]{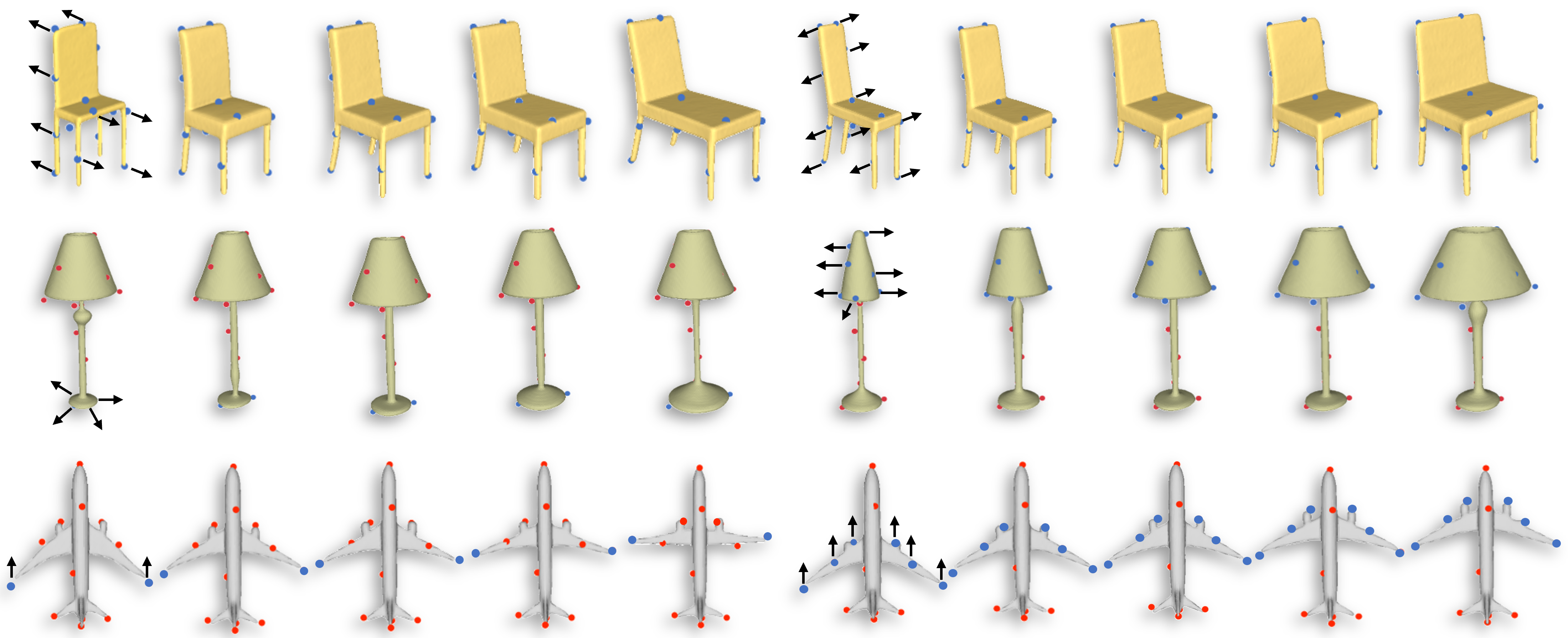}
\caption{We can use the sparse latent points to control the shape of the generated meshes. Red points are stationary, and blue points are moving. Black arrows indicate the moving direction of the blue points. Some points are invisible because they are within the mesh. 
Note that the latent points are not always strictly lying on the surface of the generated meshes. This is because our point cloud decoder assumes that some noises exist in the positions of the sparse latent points. It will generate a mesh that best fits the latent points, but avoid generating defective meshes just to strictly fit the latent points.}
\label{fig:controllable_generation}
\vspace{-1.5em}
\end{figure*}
\begin{abstract}

Mesh generation is of great value in various applications involving computer graphics and virtual content, 
yet designing generative models for meshes is challenging due to their irregular data structure and inconsistent topology of meshes in the same category.
In this work, we design a novel sparse latent point diffusion model for mesh generation.
Our key insight is to regard point clouds as an intermediate representation of meshes,
and model the distribution of point clouds instead.
While meshes can be generated from point clouds via techniques like Shape as Points (SAP),
the challenges of directly generating meshes can be effectively avoided.
To boost the efficiency and controllability of our mesh generation method,
we propose to further encode point clouds to a set of sparse latent points with point-wise semantic meaningful features,
where two DDPMs are trained in the space of sparse latent points to respectively model the distribution of the latent point positions and features at these latent points.
We find that sampling in this latent space is faster than directly sampling dense point clouds.
Moreover,
the sparse latent points also enable us to explicitly control both the overall structures and local details of the generated meshes.
Extensive experiments are conducted on the ShapeNet dataset, where our proposed sparse latent point diffusion model achieves superior performance in terms of generation quality and controllability when compared to existing methods.
Project page, code and appendix: \url{https://slide-3d.github.io}.


\end{abstract}

\vspace{-1em}
\section{Introduction}
\label{sec:intro}

Being a fundamental representation of 3D objects in computer graphics,
meshes are widely used in applications such as VR, AR, and game,
and a generative model for meshes is thus of great value.
By representing 3D objects with vertices, edges, and faces,
meshes lead to more efficient modeling and computation of object geometries.
However, such a specialized design also results in several intrinsic challenges for generative models.
The first one is the irregular data structure of meshes,
where the discrete vertex connections make it hard to define operations like convolution and up-sampling on meshes.
Moreover,
unlike point clouds and volumes, the topology of meshes is varying across different object instances in the same category.
Therefore template-based methods \cite{wang2018pixel2mesh, wen2019pixel2mesh++, gupta2020neural, liu2021deepmetahandles, yifan2020neural, jakab2021keypointdeformer} can only obtain meshes obeying the topology of used templates, 
causing defects like self-intersection when the deformation is significant.

Facing these challenges mentioned above,
we propose to generate meshes indirectly via an intermediate representation that is easier to model.
Inspired by recent successes of deep neural networks in modeling the distribution of point clouds \cite{yang2019pointflow,Luo_2021_CVPR,cai2020learning,zhou20213d,lyu2021conditional} and reconstructing meshes from point clouds \cite{peng2021shape,gupta2020neural},
we propose to use point clouds as an intermediate representation of meshes.
Consequently,
the generation of meshes is effectively reformulated as the generation of point clouds, followed by transforming point clouds into meshes.
Such a reformulation not only enables us to take advantage of the advances of point cloud generation methods,
but also successfully bypasses the aforementioned challenges,
as the distribution of point clouds is continuous and point clouds are unordered sets without explicit topology. 
In this paper, we adopt denoising diffusion probabilistic models (DDPMs)~\cite{sohl2015deep,ho2020denoising}, demonstrated promising results in modeling point clouds~\cite{Luo_2021_CVPR,zhou20213d,lyu2021conditional}, to learn the distribution of the point clouds.
And Shape as Points (SAP)~\cite{peng2021shape} is employed to reconstruct meshes from the generated point clouds, which is a powerful surface reconstruction technique that can extract high-quality watertight meshes from point clouds at low inference times.

With the introduction of point clouds as the intermediate representation of meshes, 
we can use DDPMs to model the distribution of meshes.
However, 
to ensure the quality of transformed meshes,
the generated point clouds need to be sufficiently dense.
This inevitably leads to two issues.
At first, the overall computational complexity is high,
since sampling thousands of points from DDPMs is quite time-consuming.
It is also difficult to explicitly control the structure of meshes via dense point clouds,
as the semantics of their points are not sufficiently compact.
We therefore further encode a point cloud to a sparse set of semantic latent points with features attached to every point,
and learn a \textbf{S}parse \textbf{L}atent po\textbf{I}nt \textbf{D}iffusion mod\textbf{E}l (\textbf{SLIDE})  for mesh generation following the framework of latent diffusion models~ \cite{vahdat2021score,rombach2022high}.
Specifically, 
we train two DDPMs to learn the distribution of this latent space. The first DDPM learns the distribution of positions of the sparse latent points, and the second one learns the  distribution of the features conditioned on the positions of the points.
By cascading these two DDPMs together,
we can perform unconditional generation of sparse latent points and their features.
We adopt farthest point sampling (FPS) to obtain the positions of sparse latent points from a dense point cloud,
and a neural encoder is deployed to attach each latent point a semantically meaningful feature.
Accordingly, a neural decoder is used to recover a dense point cloud from the positions and features of the sparse latent points and then reconstruct the mesh.
In this way, 
we maintain the quality of generated meshes while being able to control their overall structures and local details respectively via controlling configurations and semantic features of the sparse latent points,
as shown in Figure~\ref{fig:controllable_generation}.
Moreover,
we find that sampling in this sparse latent point space is significantly faster than directly sampling dense point clouds.
We conduct experiments on the ShapeNet~\cite{chang2015shapenet} dataset to compare both point cloud and mesh generation performance of SLIDE with other methods. 
SLIDE achieves superior performances in terms of both visual quality and quantitative metrics.
It also demonstrates great flexibility in controlling the overall structures and local part shapes of the generated objects without using any part-annotated 3D data.
In summary, the main contributions of our work are:
\textbf{1)} We propose to use point clouds as the intermediate representation of meshes.
By generating point clouds first and then reconstructing surface from them, we can generate meshes with diverse topology and high quality.
\textbf{2)} We design a novel point cloud autoencoder to further encode point clouds to a sparse set of latent points with features attached to them.
Sampling in this latent space is more efficient than directly sampling dense point clouds.
\textbf{3)} By decomposing the learning of the positions of the sparse latent points and features of them, we can perform both unconditional point cloud generation and controllable point cloud generation based on the positions of the sparse latent points as shown in Figure~\ref{fig:controllable_generation}. 
We can also perform both global and local interpolations in this latent space.

\vspace{-1em}
\section{Background}
\label{sec:back}
\begin{figure*}[t]
\vspace{-2em}
    \hspace{3em}
    \begin{subfigure}{0.9\textwidth}
    \centering
    \includegraphics[width=1\textwidth]{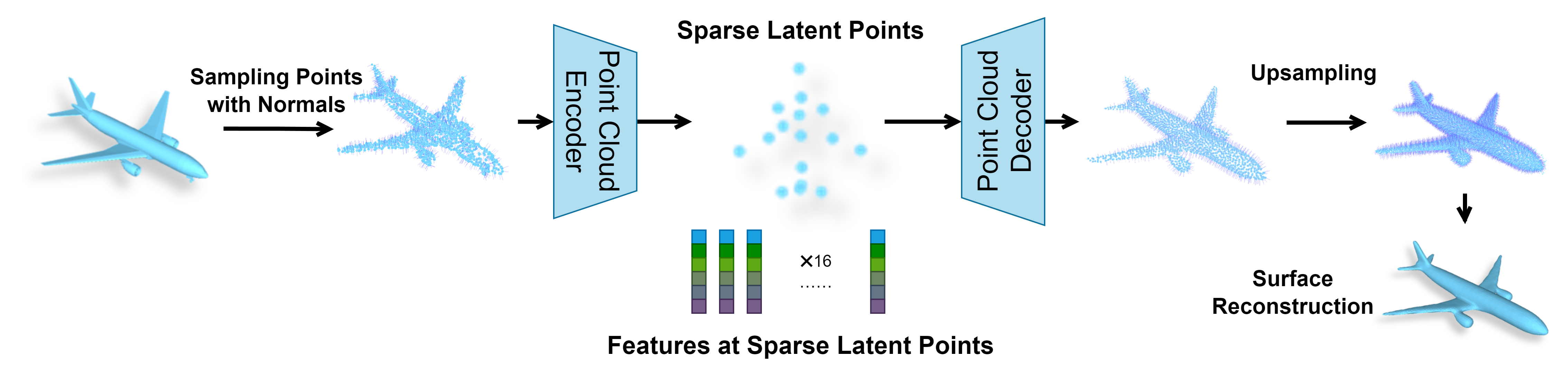}
    \caption{The autoencoder encodes a mesh to features at the sparse latent points and decodes it back to a mesh.}
    \label{fig:autoencoder_overview}
    \end{subfigure}
    
    \begin{subfigure}{0.56\textwidth}
    \centering
    \includegraphics[width=0.8\textwidth]{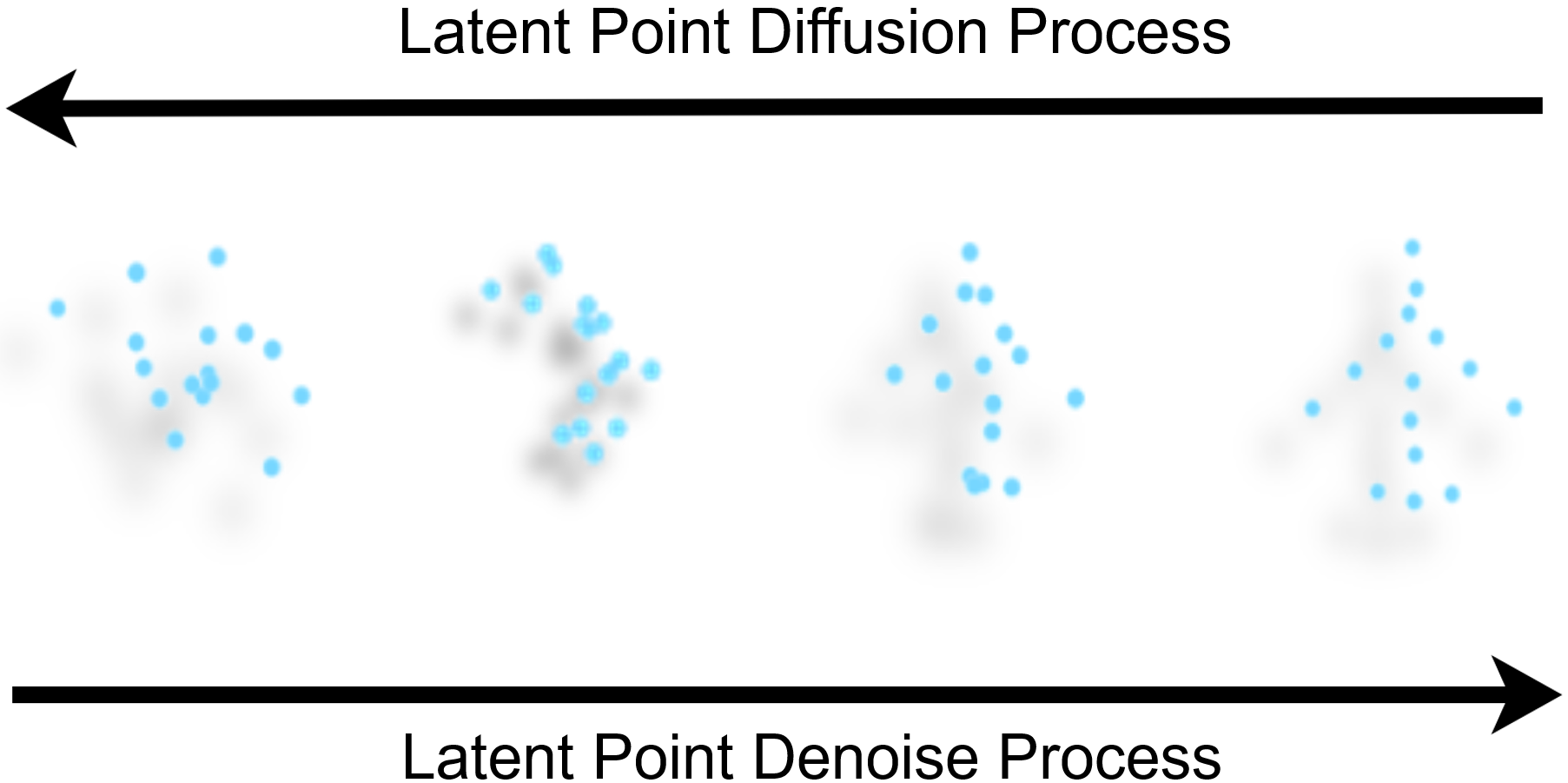}
    \caption{The DDPM learns the distribution of the sparse latent points.}
    \label{fig:latent_point_diffusion}
    \end{subfigure}
    \hspace{1em}
    \begin{subfigure}{0.42\textwidth}
    \centering
    \includegraphics[width=0.8\textwidth]{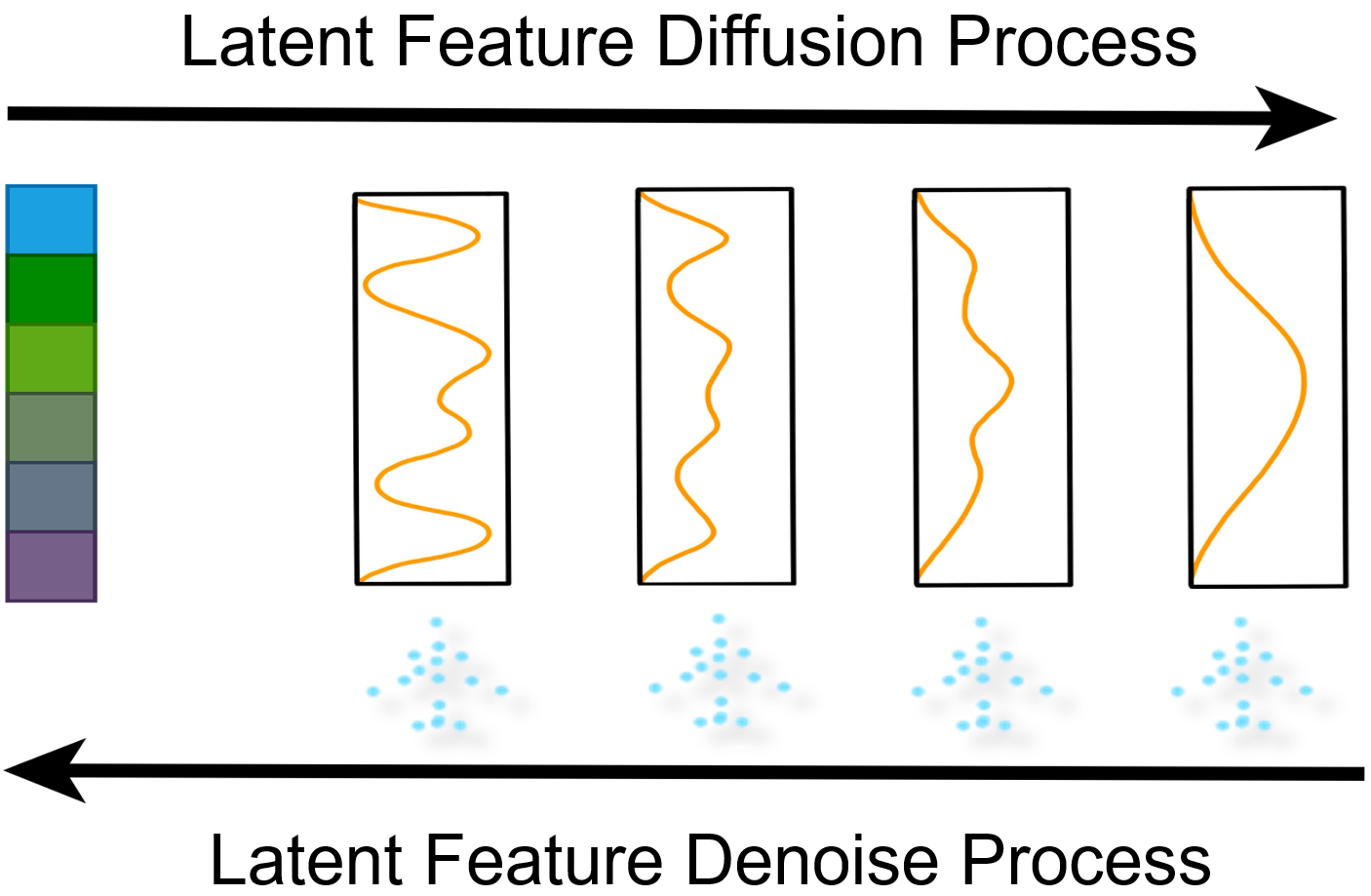}
    \caption{The DDPM learns the distribution of features at latent points.}
    \label{fig:latent_feature_diffusion}
    \end{subfigure}
    \caption{Overview of our sparse latent point diffusion model and the two latent diffusion models.}
    \vspace{-1.5em}
    \label{fig:latent_ddpm}
\end{figure*}

\subsection{Denoising Diffusion Probabilistic Model}
\label{sec:ddpm}
Denoising Diffusion Probabilistic Models (DDPMs) are a kind of generative models that learn the distribution of samples in a given dataset.
A DDPM consists of two processes: A diffusion process and a reverse process. 
The diffusion process is defined as 
\begin{align}
\label{eqn:diffusion_process}
    q(\vx^1,\cdots,\vx^T|\vx^0) = \prod_{t=1}^T q(\vx^t|\vx^{t-1}), \\
    \text{ where }
    q(\vx^t|\vx^{t-1})=\gN(\vx^t;\sqrt{1-\beta_t}\vx^{t-1},\beta_t \mI),
\end{align}
$\vx^0$ is a clean sample from the dataset, $\vx^1,\cdots,\vx^T$ are latent variables, $T$ is the number of diffusion steps,
$\gN$ denotes the Gaussian distribution and $\beta_t$'s are predefined small positive constants.
See Appendix A.4 for details of these hyper-parameters.
The diffusion process gradually adds noise to the clean sample $\vx^0$ and eventually turns it into a Gaussian noise $\vx^T$ given that $T$ is large enough.
The reverse process is defined as 
\begin{align}
\label{eqn:reverse_process}
\begin{split}
    p_{\bm{\theta}}(\vx^0,\cdots,\vx^{T-1}|\vx^T)=\prod_{t=1}^T p_{\bm{\theta}}(\vx^{t-1}|\vx^t), \\
    \text{ where }
    p_{\bm{\theta}}(\vx^{t-1}|\vx^t) = \gN(\vx^{t-1};\bm{\mu}_{\bm{\theta}}(\vx^t, t), \sigma_t^2\mI),
\end{split}
\end{align}
and the mean $\bm{\mu}_{\bm{\theta}}(\vx^t, t)$ is parameterized by a neural network.
We use the proposed method in~\cite{ho2020denoising} to reparameterize $\bm{\mu}_{\bm{\theta}}(\vx^t, t)$ as 
\begin{align}
\bm{\mu}_{\bm{\theta}}(\vx^t, t) = \frac{1}{\sqrt{\alpha_t}}\left(\vx^t-\frac{\beta_t}{\sqrt{1-\bar{\alpha}_t}}\bm{\epsilon}_{\bm{\theta}}(\vx^t, t)\right),
\end{align}
where $\alpha_t = 1 - \beta_t$, $\bar{\alpha}_t = \prod_{i=1}^t\alpha_i$.
The reverse process simulates the reverse process of the diffusion process:
It iteratively uses the network $\bm{\epsilon}_{\bm{\theta}}(\vx^t, t)$ to denoise a Gaussian noise and turn it into a clean sample.
To generate a sample from the DDPM, we first sample $\vx^T$ from the Gaussian distribution, then use Equation~\ref{eqn:reverse_process} to iteratively sample $\vx^{T-1}, \vx^{T-2}, \dots, \vx^0$ and finally obtain the sample $\vx^0$.

We use the simplified loss proposed in~\cite{ho2020denoising} to train the DDPM:
\begin{align}
\label{eqn:training objective}
    L(\bm{\theta}) = 
    \mathbb{E}_{\vx^0 \sim p_{\text{data}}}\ 
    \|\bm{\epsilon} - \bm{\epsilon}_{\bm{\theta}}(\sqrt{\bar{\alpha}_t}\vx^0 + \sqrt{1-\bar{\alpha}_t}\bm{\epsilon}, t)\|^2,
\end{align}
where $p_{\text{data}}$ is the distribution of the dataset, $t$ is sampled uniformly from $1,2,\dots,T$, and $\bm{\epsilon}$ is sampled from a Gaussian noise.
We can see that the network $\bm{\epsilon}_{\bm{\theta}}$ actually learns to predict the noise $\bm{\epsilon}$ added to the clean sample $\vx^0$.
In other words, the network $\bm{\epsilon}_{\bm{\theta}}$ learns to denoise noisy samples.
The architecture of the denoising network $\bm{\epsilon}_{\bm{\theta}}$ depends on the data format of $\vx^0$.  
If $\vx^0$ are images, a common choice of $\bm{\epsilon}_{\bm{\theta}}$ is an Unet that predicts a per-pixel adjustment of the input noisy image $\vx^t$.
If $\vx^0$ are point clouds, we can choose Point-Voxel CNN~\cite{zhou20213d} or PointNet++~\cite{lyu2021conditional} for $\bm{\epsilon}_{\bm{\theta}}$ that can predict a per-point displacement of the noisy input point cloud $\vx^t$.


\subsection{Latent Diffusion Model}
Latent diffusion models~\cite{vahdat2021score,rombach2022high} are proposed for high-resolution image synthesis.
Directly training DDPMs on high-resolution images and sampling from them are quite time-consuming.
Latent diffusion models circumvent this problem by first encoding a high-resolution image to a low-dimensional latent space, and then training DDPMs in this latent space.
Samples generated in this latent space are then decoded back to images.
We follow the same procedures as~\cite{rombach2022high} to train latent diffusion models.
First, train an autoencoder in the data space.
Then, train a DDPM using encoded samples from the dataset, namely, we can regard the variable $\vx^0$ in Section~\ref{sec:ddpm} as the encoded variable of the original data sample by the pre-trained autoencoder.


\section{Sparse Latent Point Diffusion Models}
\label{sec:Method}
\newcommand{\PositionNet}{\text{PositionNet }}
\newcommand{\conditioner}{\text{condition }}
\newcommand{\FeatureTransfer}{\text{Feature Transfer }}
\newcommand{\FT}{\text{FT }}
\newcommand{\FeatureExtractionNet}{Condition Feature Extraction subnet}
\newcommand{\DenoiseNet}{Denoise subnet}

It is difficult to directly train a generative model on meshes, because meshes have irregular data structures. In general, a mesh is composed of vertices and faces. Vertices are points in the 3D space, while faces characterize the connections among vertices.
It is easy for a generative model to model the positions of vertices, but it is difficult to model the connections among vertices.
To tackle this problem, 
we propose to use point clouds with normals as an intermediate representation of meshes for their simple structure and efficient representation.
Point clouds with normals can be sampled from the surface of meshes ($2048$ points in our experiments).
Then we can use existing generative models~\cite{luo2021diffusion, zhou20213d, DBLP:conf/eccv/CaiYAHBSH20} to model the distribution of point clouds. 
Finally, we use SAP~\cite{peng2021shape} to reconstruct meshes from the generated point clouds.
SAP is composed of an upsampling network and a Differentiable Poisson Surface Reconstruction (DPSR) algorithm.
We refer readers to the original work or Appendix A.1 for details of SAP.
As mentioned above, we can use point clouds with normals as the intermediate representation of meshes.
However, we think that point clouds are still a redundant representation of 3D shapes. 
In addition, point clouds are difficult to manipulate and control. 
To this end, we propose to further encode point clouds with normals to some sparse latent points with features as shown in Figure~\ref{fig:autoencoder_overview}.
The intuition behind this representation is that a 3D shape can be decomposed to its skeleton that encodes the overall structure of the shape, and features located on the skeleton that encodes the local geometric details of the shape.
To make the sparse latent points stretch over a given point cloud, we use farthest point sampling (FPS) to sample a given number ($16$ in our experiments) of points as the sparse latent points.
We use two strategies to choose the first point in FPS: The first is choosing the centroid (mean coordinates of all the points) as the first point in FPS. The second is randomly choosing a point as the first point.
We then design a point cloud encoder to encode a point cloud with normals to features attached to the sampled sparse latent points. 
We also design a point cloud decoder to decode the sparse latent points with features back to the input point cloud with normals.
The details of the point cloud encoder and decoder are explained in the next section.



\begin{figure*}[t]
    \vspace{-2em}
    \begin{subfigure}{0.53\textwidth}
    \centering
    \includegraphics[width=0.95\textwidth]{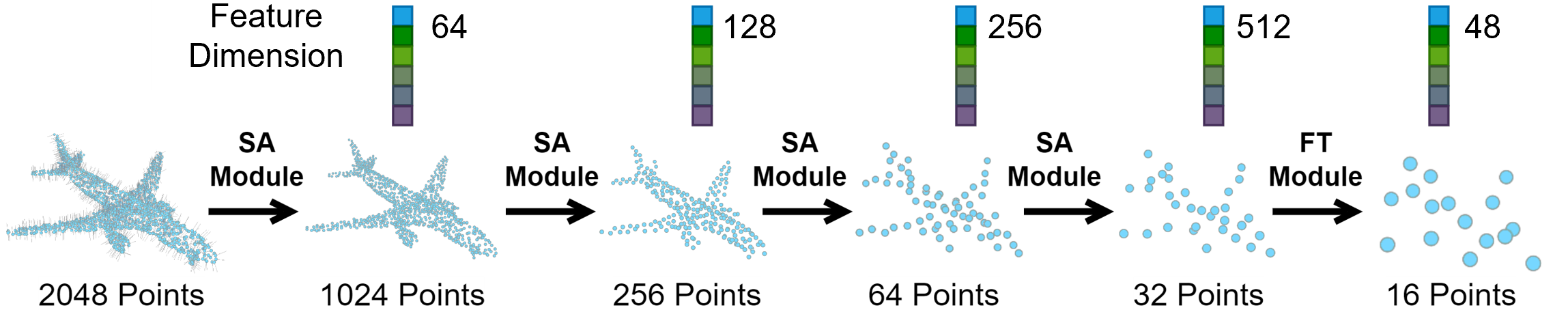}
    \caption{The point cloud encoder.}
    \label{fig:pcd_encoder}
    \end{subfigure}
    \begin{subfigure}{0.46\textwidth}
    \centering
    \includegraphics[width=0.95\textwidth]{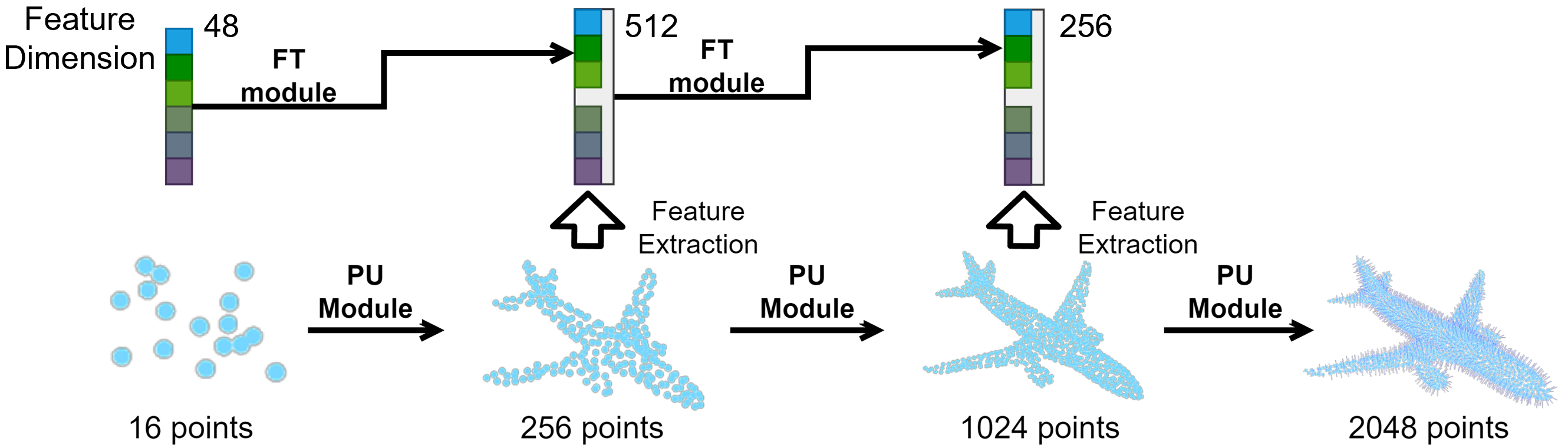}
    \caption{The point cloud decoder.}
    \label{fig:pcd_decoder}
    \end{subfigure}
    \vspace{-0.7em}
    \caption{Architecture of the point cloud autoencoder.}
    \vspace{-1.5em}
    \label{fig:pcd_autoencoder}
\end{figure*}
\subsection{Architecture of the autoencoder}
As mentioned above, we need a point cloud encoder to encode a point cloud to a sparse set of points with features, and a decoder to decode the sparse latent points back to the input point cloud. 
In this section, we explain the detailed architectures of the point cloud encoder and decoder.
\vspace{-1.5em}
\paragraph{Point cloud encoder.}
The encoder needs to encode a point cloud to features at the FPS sampled sparse set of points.
The overview of the encoder is shown in Figure~\ref{fig:pcd_encoder}.
It mainly consists of the improved Set Abstraction (SA) modules with attention mechanism proposed in PDR~\cite{lyu2021conditional}.
We briefly repeat the design of the SA module.
The input of the SA module is a set of points with a feature attached to each point. The SA module subsamples the input points by farthest point sampling (FPS) and then propagates features to the subsampled points. 
Specifically, for every point in the subsampled points, it finds $K$ nearest neighbors in the input points. 
Then it transforms the features of the neighbors by a shared Multi-layer perceptron (MLP) and aggregates the transformed features to this point through the attention mechanism. 
We refer readers to the original work~\cite{lyu2021conditional} for details of the SA module.
In our encoder, there are $4$ cascaded SA modules. They iteratively downsample the input point cloud ($2048$ points) to $1024, 256, 64, 32$ points and propagate features to the downsampled points. The features of the input point cloud are simply the 3D coordinates and normals of every point.

Recall that the encoder needs to encode the input point cloud to features at the sampled sparse latent points ($16$ points). 
We achieve this by mapping features at the output of the last level SA module (consisting of $32$ points) to the sparse latent points.
We use the feature transfer (FT) module proposed in PDR~\cite{lyu2021conditional} to map features from the last level SA module to the sparse latent points.
The FT module can map features from one set of points to the second set of points, and the mapping is parameterized by a neural network. 
It is worth noting that the FT module requires points in the second set to have features, and these features are used as queries to aggregate features from the first set of points to the second set.
We refer readers to Appendix A.2 or the original work~\cite{lyu2021conditional} for details of the FT module.
To utilize the FT module, we first use a lightweight PointNet++ to extract features for the sparse latent points themselves.
Then we use it to map features from the last level SA module to the sparse latent points. 
The mapped features are concatenated to the original features at the sparse latent points to form the final features that represent the input point cloud.

Overall, as shown in Figure~\ref{fig:pcd_encoder}, our point cloud encoder contains $4$ SA modules to hierarchically extract features from the input point cloud, a light-weight PointNet++ to extract features for the sparse latent points, and an FT module to map features from the last level SA module to the sparse latent points. 


\vspace{-1.5em}
\paragraph{Point cloud decoder.}
As mentioned above, the point cloud encoder can encode an input point cloud to features at the sampled sparse latent points.
Next, we explain the point cloud decoder that can decode the sparse latent points with features back to the input point cloud with normals.
Its overall structure is shown in Figure~\ref{fig:pcd_decoder}.
It contains $3$ point upsampling (PU) modules that gradually upsample the sparse latent points ($16$ points) to $2048$ points.

The input to the $l$-th PU module is a set of points $\mX^l = \{x_{j}^l \in \mathbb{R}^{3} | 1 \leq j \leq N^{l} \}$, with features attached to each point $\mF^{l} = \{f_{j}^{l} \in \mathbb{R}^{d^{l}} | 1 \leq j \leq N^{l} \}$, where $N^{l}$ is the number of input points to the $l$-th PU module, $f_j^{l}$ is the feature at point $x_j^{l}$, and $d^{l}$ is the dimension of the feature.
The input to the first PU module is the sparse latent points $\mX^1$ and their features $\mF^1$.
The PU module first uses a shared Multi-layer Percpetron (MLP) to transform the feature $f_j^{l}$ at every point $x_j^{l}$ to $\gamma$ displacements. 
Then the displacements are added to the original point  $x_j^{l}$ to obtain $\gamma$ new points. 
In this way, the input points are upsampled by a factor of $\gamma$.
To enforce the uniformness of the upsampled points, after upsampling, we use FPS to downsample the upsampled points by half.
Overall, the input points are upsampled by a factor of $\gamma/2$ and we obtain the upsampled points $\mX^{l+1}=\{x_j^{l+1} \in \mathbb{R}^{3} | 1 \leq j \leq N_{l+1} \}$, where $N_{l+1} = \gamma N_{l}/2$.
Note that the upsampled points $\mX^{l+1}$ are controlled by the features $\mF^{l}$ at the input points $\mX^{l}$ and the learned MLP.

Next, we use the $(l+1)$-th PU module to further upsample the output of the $l$-th PU module, $\mX^{l+1}$, but first we need to compute features for points in $\mX^{l+1}$, because the PU module needs features at the input points to perform upsampling.
We think that the features should consist of two parts. The first part is from the upsampled points $\mX^{l+1}$ themselves. 
This part of the feature characterizes the shape of the current point cloud $\mX^{l+1}$, and instructs how we can further upsample and refine it to make it more plausible.
We use a improved PointNet++~\cite{lyu2021conditional} to extract this part of the feature from the set $\mX^{l+1}$, and denote it as $\mF_1^{l+1} = \{f_{1,j}^{l+1} \in \mathbb{R}^{d_1^{l+1}} | 1 \leq j \leq N_{l+1} \}$, where $f_{1,j}^{l+1}$ is the feature at point $x_j^{l+1}$, and $d_1^{l+1}$ is the dimension of the feature.

The second part of the feature should come from the previous level PU module, namely, $\mX^{l}$.
This is because we want the information in the features at the sparse latent points can propagate along the PU modules layer by layer to control the shape of the final decoded point cloud.
Afterall, all information of the input point cloud is encoded to the features at the sparse latent points, and we want the decoded shape to be consistent with the information stored in the features at the sparse latent points.
Therefore, to obtain the second part of the feature for every point in $\mX^{l+1}$, we use the FT module mentioned in the point cloud encoder to map features from $\mX^{l}$ to $\mX^{l+1}$, and the first part feature $\mF_1^{l+1}$ at $\mX^{l+1}$ are used as queries.
After obtaining the second part of the features, $\mF_2^{l+1} = \{f_{2,j}^{l+1} \in \mathbb{R}^{d_2^{l+1}} | 1 \leq j \leq N_{l+1} \}$, it is concatenated with the first part feature $\mF_1^{l+1}$ to obtain the final feature for $\mX^{l+1}$: $\mF^{l+1} = \{(f_{1,j}^{l+1}, f_{2,j}^{l+1}) \in \mathbb{R}^{d^{l+1}} | 1 \leq j \leq N_{l+1} \}$, where $d^{l+1} = d_1^{l+1} + d_2^{l+1}$. 

After upsampling the point cloud $\mX^{l}$ to $\mX^{l+1}$, and obtaining the features $\mF^{l+1}$ at $\mX^{l+1}$, we can use the PU module to further upsample $\mX^{l+1}$.
By applying the PU module and FT module iteratively, we can gradually upsample the sparse latent points to a point cloud of $2048$ points.
For the last PU module, we let it predict both $\gamma$ displacements and $\gamma$ normals, so that the final output point cloud has $2048$ points with normals.
Overall, the input to the point cloud decoder is the sparse latent points $\mX^{1}$ ($16$ points) and their features $\mF^{1}$.
The decoder outputs the intermediate results $\mX^{2}$ ($256$ points), $\mX^{3}$ ($1024$ points), the final reconstructed point cloud $\mX^{4}$ ($2048$ points) and normals $\mF^{4}$.
\vspace{-1.5em}
\paragraph{Training of the autoencoder.}
The point cloud autoencoder is trained to encode the input point cloud and then reconstruct the point cloud.
The input to the autoencoder is point cloud $\mX_{\text{in}}$ ($2048$ points) with normals $\mF_{\text{in}}$ sampled from the meshes in the dataset.
The supervision is added on all the intermediate upsampling results in the point cloud decoder: $\mX^{2}, \mX^{3}$, $\mX^{4}$.
The loss is the sum of the Chamfer distance (CD) between $\mX_{\text{in}}$ and $\mX^{2}, \mX^{3}$, $\mX^{4}$, respectively.
Note that when computing the CD loss between $\mX_{\text{in}}$ and $\mX^{2}, \mX^{3}$, we first downsample $\mX_{\text{in}}$ using farthest point sampling to the same number of points as $\mX^{2}$ and $\mX^{3}$, respectively.
We also add a normal consistency loss between the ground-truth normals $\mF_{\text{in}}$ and the predicted normals $\mF^4$ with a weight of $0.1$. See Appendix B.3 for details of this loss.
We further add a slight Kullback–Leibler divergence loss (weight $10^{-5}$) between the encoded features $\mF^{1}$ and a standard normal distribution.
This regularization term is to encourage the latent feature space to be simple and smooth, so that we can perform manipulation and interpolation in this space.
Before the encoder encodes the input point cloud $\mX_{\text{in}}$ to the sampled sparse latent points $\mX^{1}$, we add a Gaussian noise with a standard deviation of $0.04$ to the point positions in $\mX^{1}$.
This is to make the autoencoder more robust to the positions of the sparse latent points, so that even if the positions of the sparse latent points are not perfect (\eg, human-edited sparse latent points), the autoencoder can still well reconstruct the input point cloud.

\subsection{Train DDPMs in the Sparse Latent Point Space}
\label{sec:train_latent_ddpm}
After training the autoencoder, we can train latent DDPMs in the latent space of the autoencoder, while freezing the parameters of the autoencoder.
Specifically, for each point cloud, we can encode it to features $\mF^{1}$ at sparse latent points $\mX^{1}$.
We train two DDPMs in this latent space. The first one learns the distribution of the sparse latent points $\mX^{1}$ and is illustrated in Figure~\ref{fig:latent_point_diffusion}.
The sparse latent points can be seen as a point cloud with very few points. Therefore, its distribution can be effectively learned by a DDPM designed for point clouds, except that we can use a light-weight PointNet++ as the denoising network $\bm{\epsilon}_{\bm{\theta}}(\vx^t, t)$ in Equation~\ref{eqn:training objective}.

The second DDPM learns the distribution of the feature $\mF^{1}$ given the sparse latent points $\mX^{1}$, which is illustrated in Figure~\ref{fig:latent_feature_diffusion}.
For algebraic simplicity, we use $\vf$ to denote the vector form of $\mF^{1}$, namely, concatenating all feature vectors in $\mF^{1}$ to a single  vector $\vf$. Similarly, let $\vx$ be the vector form of $\mX^{1}$.
The second DDPM can be seen as a conditional DDPM that generates features $\vf$ conditioned on the positions of the sparse latent points $\vx$.
To achieve this, we can simply replace the diffusion variable $\vx$ in Equation~\ref{eqn:diffusion_process} with $\vf$, and adapt the reverse process in Equation~\ref{eqn:reverse_process} to
\vspace{-1em}
\begin{align}
\begin{split}
    p_{\bm{\phi}}(\vf^{0},\cdots,\vf^{T-1}|\vf^{T},\vx)&=\prod_{t=1}^T p_{\bm{\phi}}(\vf^{t-1}|\vf^{t}, \vx), \\
    \text{ where }
    p_{\bm{\phi}}(\vf^{t-1}|\vf^{t}, \vx) &= \gN(\vf^{t-1};\bm{\mu}_{\bm{\phi}}(\vf^{t}, \vx, t), \sigma_t^2\mI),
\end{split}
\end{align}
and $\bm{\mu}_{\bm{\phi}}(\vf^{t}, \vx, t)$ is parameterized as 
\begin{align}
\bm{\mu}_{\bm{\phi}}(\vf^{t}, \vx, t) = \frac{1}{\sqrt{\alpha_t}}\left(\vf^{t}-\frac{\beta_t}{\sqrt{1-\bar{\alpha}_t}}\bm{\epsilon}_{\bm{\phi}}(\vf^{t}, \vx, t)\right).
\end{align}
To feed the input $\vf^{t}$ and $\vx$ to the denoising network $\bm{\epsilon}_{\bm{\phi}}$, we concatenate each feature in $\vf^{t}$ to the corresponding point in $\vx$.
In other words, the input to $\bm{\epsilon}_{\bm{\phi}}$ can be seen as a sparse point cloud with noisy features attached to each point, and the output of the network $\bm{\epsilon}_{\bm{\phi}}$ is to predict the noise added to each feature in $\vf$.
Therefore, we can use the improved PointNet++ in PDR~\cite{lyu2021conditional} as the denoising network $\bm{\epsilon}_{\bm{\phi}}$.
Correspondingly, the training loss of the network $\bm{\epsilon}_{\bm{\phi}}$ is 
\begin{align}
    L(\bm{\phi}) = 
    \mathbb{E}_{(\mX_{\text{in}}, \mF_{\text{in}}) \sim p_{\text{data}}}\ 
    \|\bm{\epsilon} - \bm{\epsilon}_{\bm{\phi}}(\sqrt{\bar{\alpha}_t}\vf + \sqrt{1-\bar{\alpha}_t}\bm{\epsilon}, \vx, t)\|^2, \nonumber
\end{align}
where $\mX_{\text{in}}$ is the point cloud ($2048$ points) sampled from a mesh in the dataset, $\mF_{\text{in}}$ are the corresponding normals, $\vx$ are sampled sparse latent points from $\mX_{\text{in}}$, $\vf$ is the encoded feature at $\vx$ obtained by the trained authoencoder, $t$ is sampled uniformly from $1,2,\dots,T$, and $\bm{\epsilon}$ is sampled from a Gaussian noise.

The detailed architecture of the two DDPMs is provided in Appendix A.3. 
After training the two DDPMs, we can use them to perform both unconditional 3D shape generation or controllable generation conditioned on the positions of the sparse latent points. 
To perform unconditional 3D point cloud generation, we can simply cascade the two DDPMs together: The first DDPM generates a set of sparse latent points, and the second DDPM generates features at the sparse latent points. Finally, the point cloud decoder decodes the sparse latent points with features to a point cloud.
To achieve controllable generation, we can manipulate the positions of the sparse latent points, then feed the human-adjusted sparse latent points to the second DDPM to generate plausible features on them, and finally decode them to a point cloud.



\section{Related Work}
\label{sec:related}
\begin{figure*}[t]
    \vspace{-2em}
    \centering
    \includegraphics[width=1\textwidth]{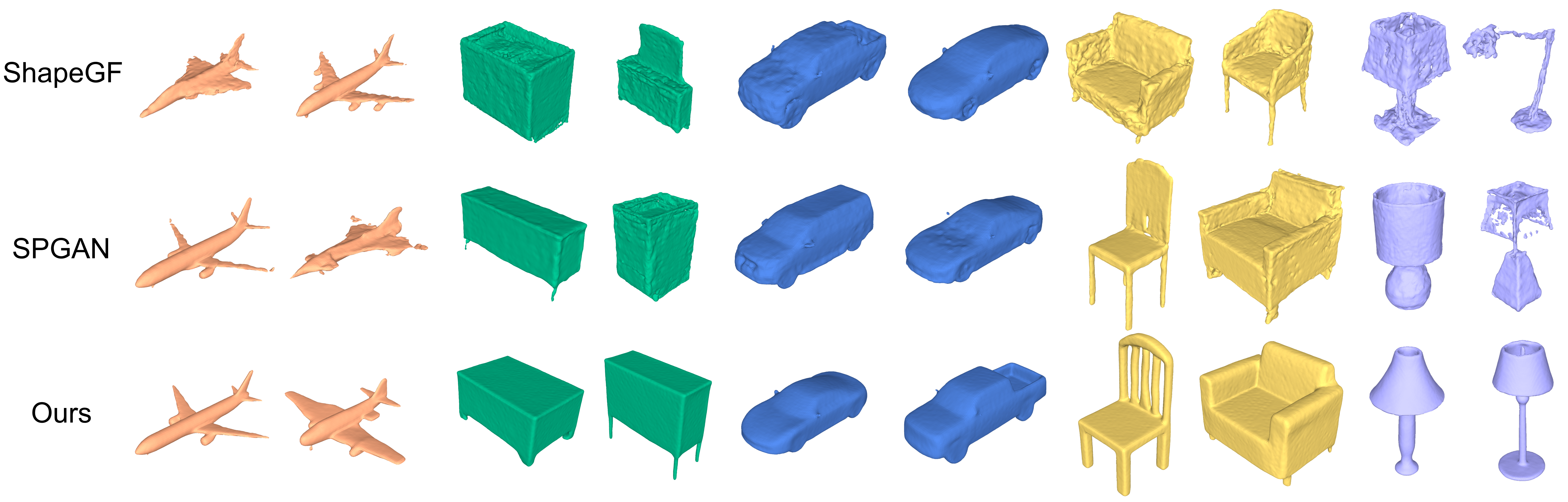}
    \vspace{-2em}
    \caption{Mesh generated by our methods and baselines. 
    We can see that meshes generated by our method are more visually appealing. More examples of other baselines and our method are provided in Appendix B.5 and B.8.}
    \label{fig:mesh_comparison}
    \vspace{-1.5em}
\end{figure*}

\begin{figure}[t]
    \centering
    \includegraphics[width=0.5\textwidth]{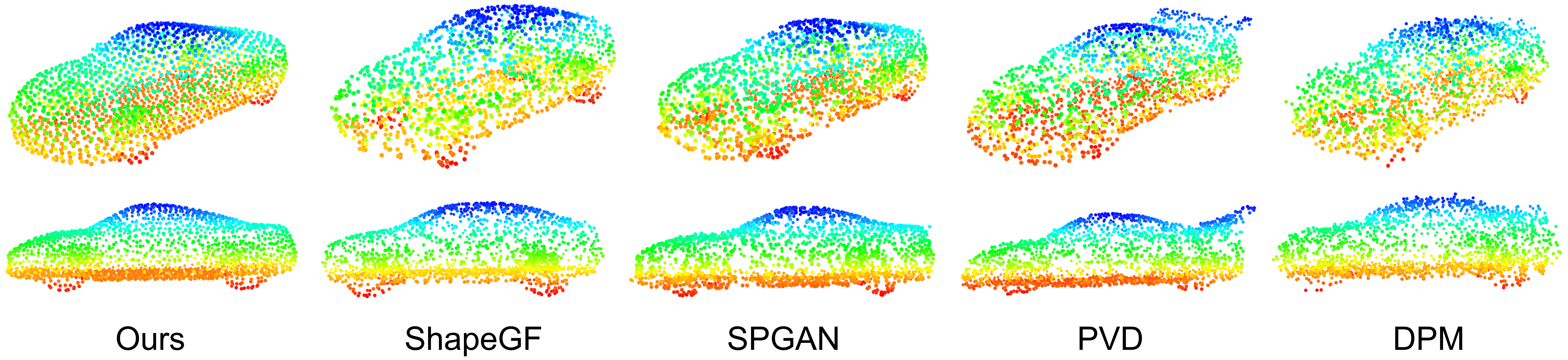}
    \caption{Point clouds generated by our method and baselines. More examples are provided in Appendix B.7 and B.8.}
    \label{fig:car_pc}
    \vspace{-1em}
\end{figure}

\paragraph{Mesh Generation.}
Most existing mesh generation methods rely on deforming a template mesh or another mesh~\cite{wang2018pixel2mesh, wen2019pixel2mesh++, gupta2020neural, liu2021deepmetahandles, yifan2020neural, jakab2021keypointdeformer, jiang2020shapeflow}, but meshes generated in this way are usually limited by the topology of the template or the initial mesh. And large deformations could cause defects. In contrast, our method is able to generate meshes from scratch with diverse topologies. 
Another line of works uses implicit representations of 3D shapes~\cite{DBLP:conf/cvpr/ChenZ19, park2019deepsdf, sitzmann2020implicit, genova2020local, zheng2022sdf, kleineberg2020adversarial,Chen_2021_ICCV, mittal2022autosdf}, but it usually requires dense neural network evaluations to extract meshes from the learned model.

\vspace{-1.5em}
\paragraph{Point cloud generation.}
Many learning-based methods are proposed to model the distribution of point clouds.
Some works use generative adversarial networks (GANs) to generate point clouds~\cite{DBLP:conf/iccv/TreeGan,li2021sp,achlioptas2018learning,li2018point}.
\cite{achlioptas2018learning} also trains a latent GAN in the latent space of a point cloud autoencoder, but the autoencoder they use can only encode a point cloud to a global feature.
Other works~\cite{yang2019pointflow,DBLP:conf/nips/SoftFlow,DBLP:conf/eccv/DPF-Net} use normalizing
flows to model the distribution of point clouds.
ShapeGF~\cite{DBLP:conf/eccv/CaiYAHBSH20} learns gradient fields to move randomly sampled points to the surface of the objects.
DDPMs have also been applied to point cloud generation~\cite{luo2021diffusion, zhou20213d}.
The generated point clouds of these methods can be transformed to meshes through surface reconstruction techniques~\cite{hanocka2020point2mesh,wei2021deep,chen2022neural,jiang2020shapeflow, williams2019deep, chibane2020implicit, gao2020learning, shen2021deep}.
In this work, we choose SAP~\cite{peng2021shape} for surface reconstruction for its efficiency and reconstruction quality.
\vspace{-1.5em}
\paragraph{Diffusion models.}
DDPMs are a kind of likelihood-based generative model that generate samples by gradually denoising a Gaussian noise~\cite{ho2020denoising,sohl2015deep}.
They have shown promising results for 3D point cloud generation~\cite{luo2021diffusion, zhou20213d}.
Our work is based on the recently proposed
latent diffusion models~\cite{rombach2022high, vahdat2021score}.
Latent diffusion models train diffusion models in the latent space of an autoencoder that encodes data samples to a more compact representation, and thus makes the training and sampling process of DDPMs faster.

\vspace{-1.5em}
\paragraph{Concurrent works.}
The concurrent work, LION~\cite{zeng2022lion}, also proposes to use a latent diffusion model to learn the distribution of point clouds and then use SAP~\cite{peng2021shape} to reconstruct meshes from point clouds, but the latent point cloud representation they use is a noisy point cloud with the same number of points ($2048$ points) as the original clean point cloud.
In contrast, we encode the original clean point cloud to a sparse set of latent points ($16$ points) with features of dimension $48$, which is a more compact representation and thus leads to faster training and sampling for DDPMs.
This representation also enables us to perform controllable generation using the sparse latent points.
Concurrently, NVMG~\cite{zheng2022neural} proposes to use voxels as the latent representations of meshes, but computational cost increases rapidly as the resolution of the 3D grid increases.

\section{Experiment}
\label{sec:experiment}

We present our main experiment results in this section. 
We use ShapeNet~\cite{chang2015shapenet} to train our mesh generative model, SLIDE, and compare it with other baselines.
We use the pre-processed ShapeNet dataset provided by~\cite{peng2021shape}.
The detailed setups and complete experiment results are provided in Appendix B. 



\subsection{Evaluation Metrics}
To evaluate the quality of generated meshes, we uniformly sample point clouds ($2048$ points) with normals from the generated meshes and reference meshes from the validation set. 
Then we use the commonly used point cloud evaluation metrics 1-NN~\cite{yang2019pointflow}, Minimum Matching Distance (MMD) and Coverage (COV) as our main evaluation tools.
All of the metrics require a distance metric to compute the distance between two point clouds.
We use the commonly used Chamfer distance (CD) and earth mover distance (EMD).
We also use the normal consistency loss between two point clouds with normals.
We find that it can better reflect the surface curvature differences between the two underlying meshes.
Details of the normal consistency loss are described in Appendix B.3.

\subsection{Point Cloud and Mesh Generation}
We train SLIDE on $5$ categories of the ShapeNet dataset: Airplane, cabinet, car, chair, and lamp.
And compare with baselines TreeGan~\cite{DBLP:conf/iccv/TreeGan}, 
SPGAN~\cite{li2021sp},
ShapeGF~\cite{cai2020learning}, PVD~\cite{zhou20213d}, 
DPM~\cite{luo2021diffusion}.
All the baselines are trained by ourselves using their public codebase.
We compare both the point clouds that they generate and meshes reconstructed from the point clouds using SAP.
Meshes generated by SLIDE and baselines are shown in Figure~\ref{fig:mesh_comparison}.
More examples and generated point clouds are shown in Appendix B.6, B.7, and B.8.
We can see that SLIDE generates meshes of the highest visual quality, with smooth surfaces and sharp details.
Since all the meshes are reconstructed from the generated point clouds using the same method, SAP.
This means the quality of the generated point clouds greatly affects the quality of the reconstructed meshes.
We provide an example of generated point clouds in Figure~\ref{fig:car_pc}. More point cloud examples are provided in Appendix B.7. 
Indeed, we can see that point clouds generated by SLIDE spread more uniformly on the surface of the objects, and bear less noise compared with other methods. 
We attribute this to the design of our novel point cloud autoencoder.
Quantitatively, we compute 1-NN, MMD, and COV on both generated point clouds and reconstructed meshes. Results are shown in Appendix B.4 and B.6.
In terms of efficiency, the average generation time for a single point cloud of SLIDE is about 0.2s (See Appendix B.10 for more details of the generation time.) tested on a single NVIDIA A100 GPU, while the DDPM-based method that directly trains generative models on dense point clouds, PVD~\cite{zhou20213d}, need 2.93s to generate a point cloud tested on the same A100 GPU.
LION~\cite{zeng2022lion} reports it needs 27.12s per shape.
We also conduct an ablation study on the number of sparse latent points and the method to sample them. Results are shown in Appendix B.10.

\begin{figure}[t]
    \centering
    \includegraphics[width=0.4\textwidth]{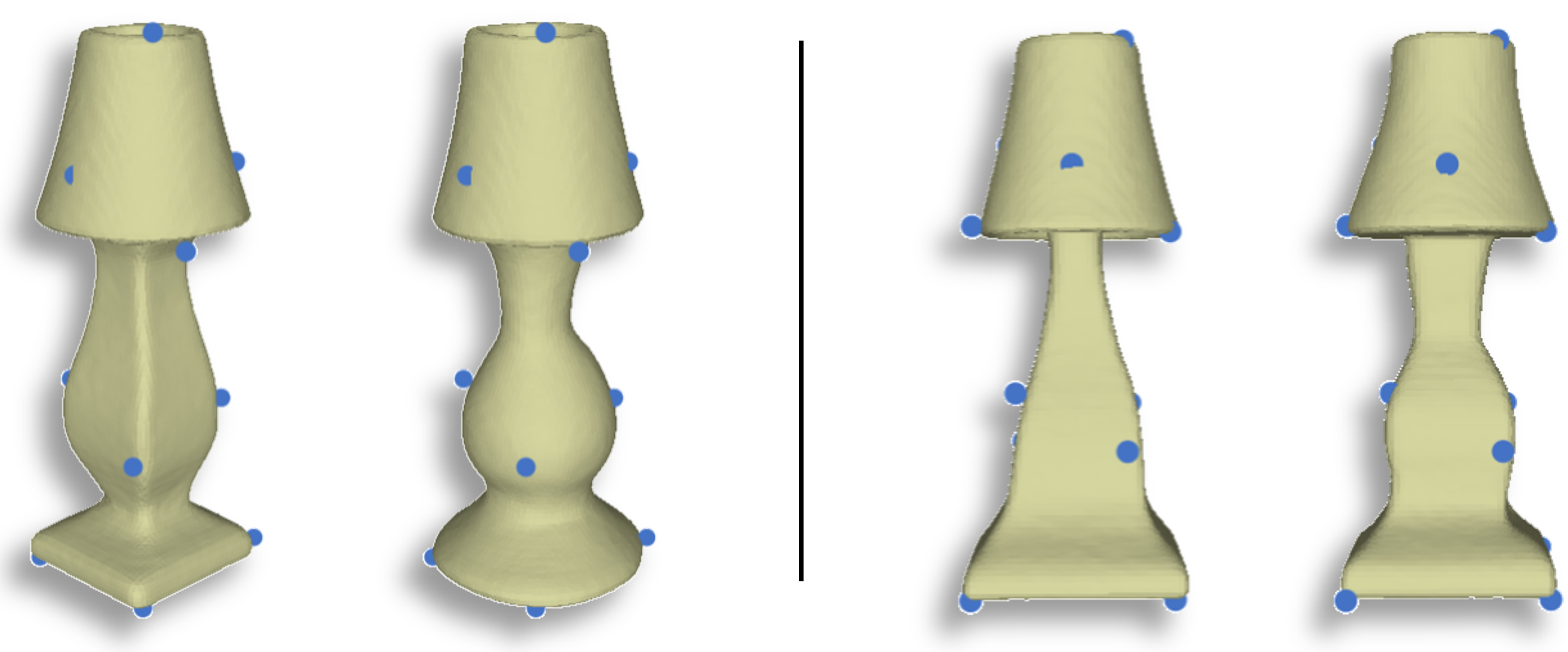}
    \vspace{-1em}
    \caption{SLIDE is able to generate diverse meshes for the same set of sparse latent points due to the stochasticity in the feature generation process. Here are two pairs of generated lamps for the same set of latent points.}
    \label{fig:generation_diversity}
\vspace{-1.5em}
\end{figure}

\begin{figure}[t]
    \centering
    \includegraphics[width=0.95\linewidth]{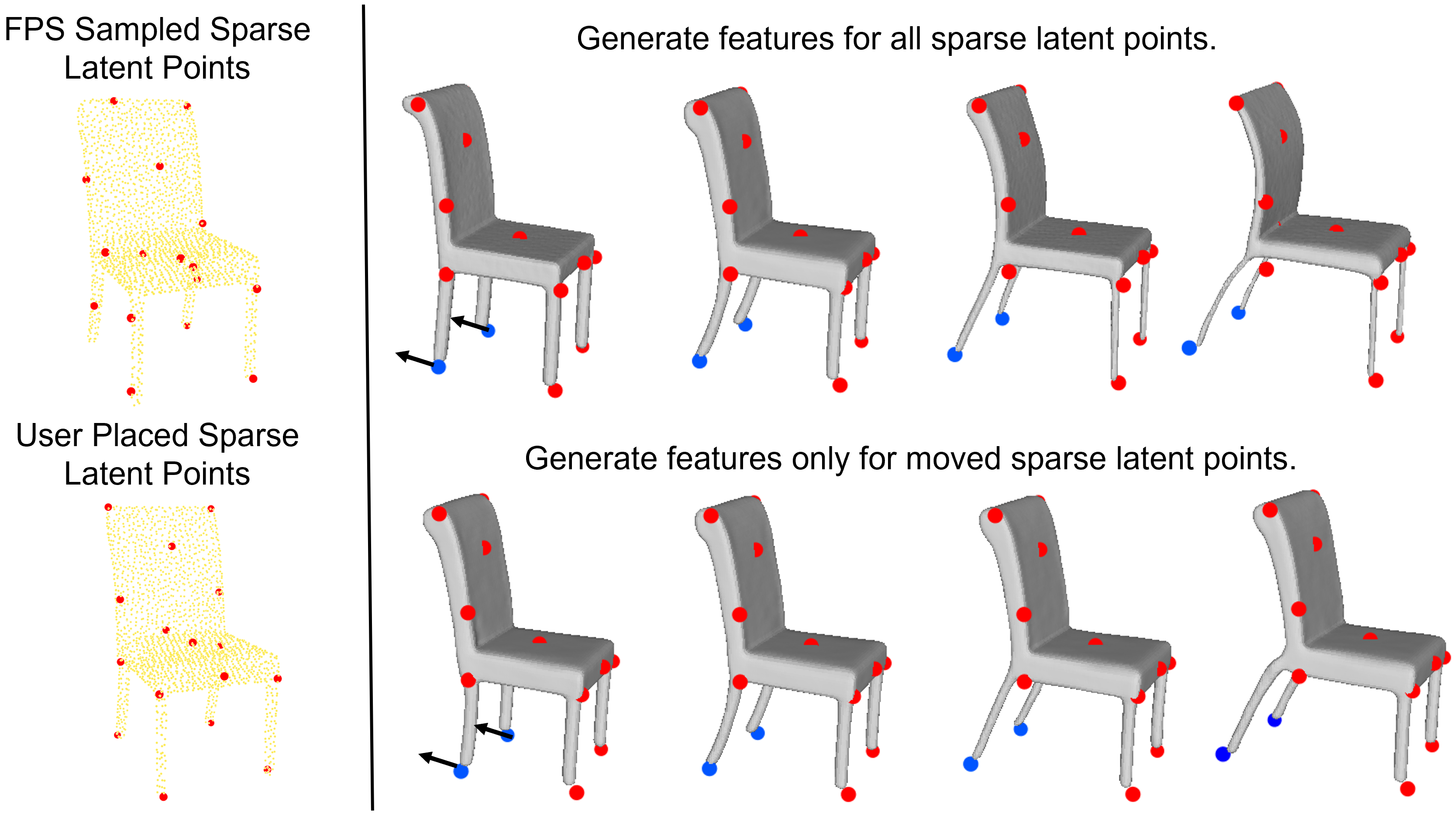}
    \vspace{-1em}
    \caption{
    Use manually placed sparse latent points to control the rear legs of the generated chairs. 
    The blue points are moving and the red points are fixed.
    The top row generates new features for all sparse latent points, and the bottom row generates new features only for moved points and fixes the features of the rest points.}
    \label{fig:local_controllable_generation}
\end{figure}

\begin{figure}[t]
\vspace{-1em}
    \centering
    \includegraphics[width=0.7\linewidth]{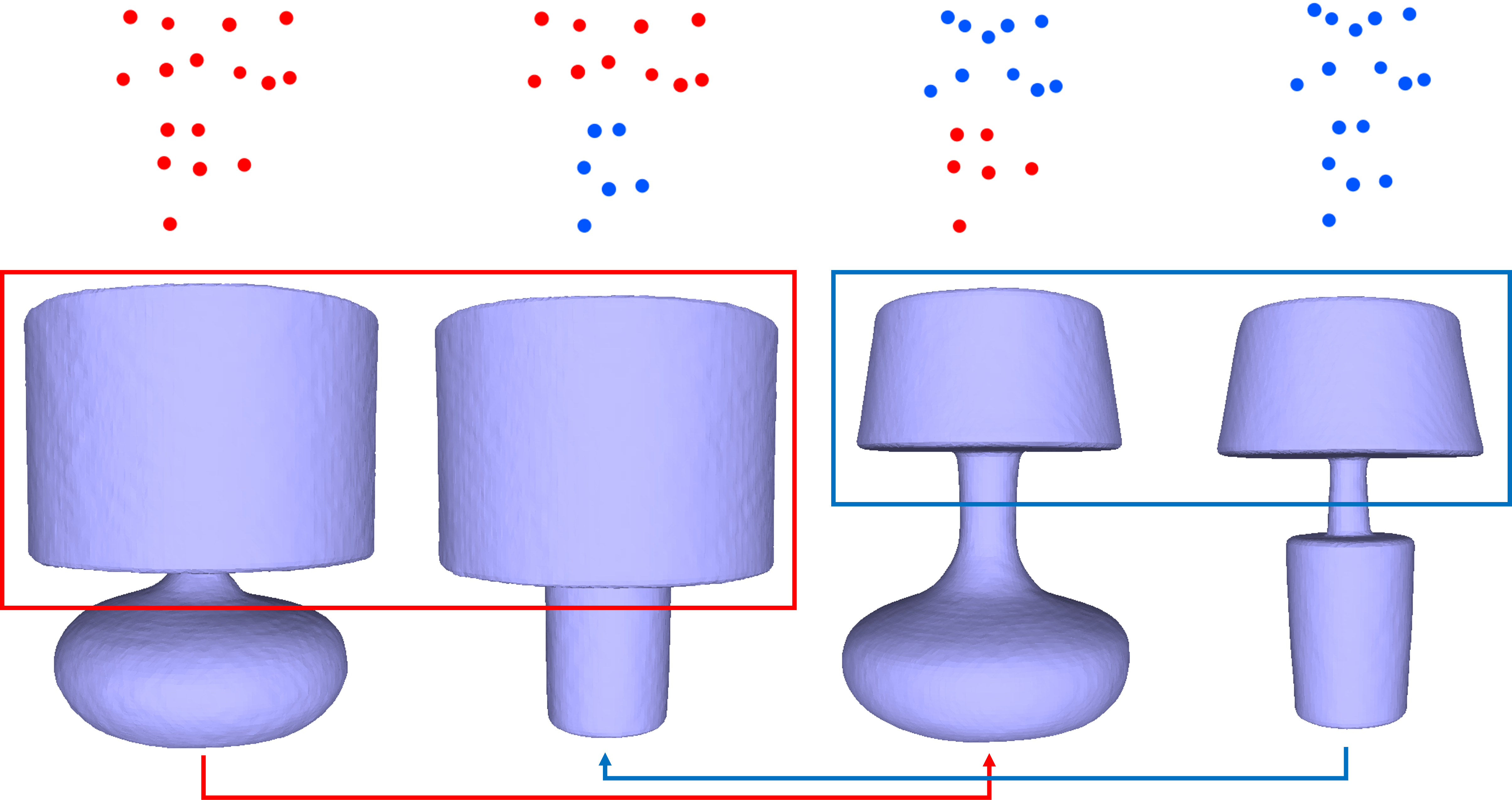}
    \vspace{-0.5em}
    \caption{Perform shape combination. 
    The first row are the sparse latent points of the original two lamps and the combined sparse latent points.
    The second row are the original two lamps (two sides) and the two lamps (middle two) obtained by combining the top part and bottom part of the original lamps.}
    \label{fig:shape_combination}
\vspace{-1.5em}
\end{figure}


\vspace{-2em}
\paragraph{Controllable Generation.}
As mentioned in Section~\ref{sec:train_latent_ddpm}, we can use the sparse latent points to control the generated mesh.
Specifically, we can change the positions of the sparse latent points, then use the second DDPM to generate features at the latent points, and finally decode them to a point cloud and reconstruct the mesh.
Several examples are shown in Figure~\ref{fig:controllable_generation}. 
It shows that we can use the sparse latent points to control the overall scale of the generated mesh as well as change the position, scale, or shape of a part of the mesh.
It is worth noting that we achieve this without any part annotations of the dataset.
SLIDE is also able to generate diverse meshes even for the same set of sparse latent points due to the stochasticity in the feature generation process.
Figure~\ref{fig:generation_diversity} gives two pairs of examples.

The sparse latent points in Figure~\ref{fig:controllable_generation} are obtained by FPS. 
At inference, we can also manually place the sparse latent points at regions of interest other than FPS sampled points and control the corresponding part.
This is because we augment the FPS sampled sparse latent points with Gaussian noises during training and it makes our model robust to the positions of the sparse latent points.
Figure~\ref{fig:local_controllable_generation} gives an example where we manually select the sparse latent points and control the rear legs of a chair.
In addition, if we want to keep the rest part of a shape fixed while changing the part we want to edit, we can use the second DDPM to sample features only for moved sparse latent points and fix the features of rest points.
See Figure~\ref{fig:local_controllable_generation} for an example.
This is achieved by an algorithm similar to DDPM-based image inpainting and is described in Appendix A.5.

\begin{figure}[t]
\vspace{-2em}
    \centering
    \includegraphics[width=0.43\textwidth]{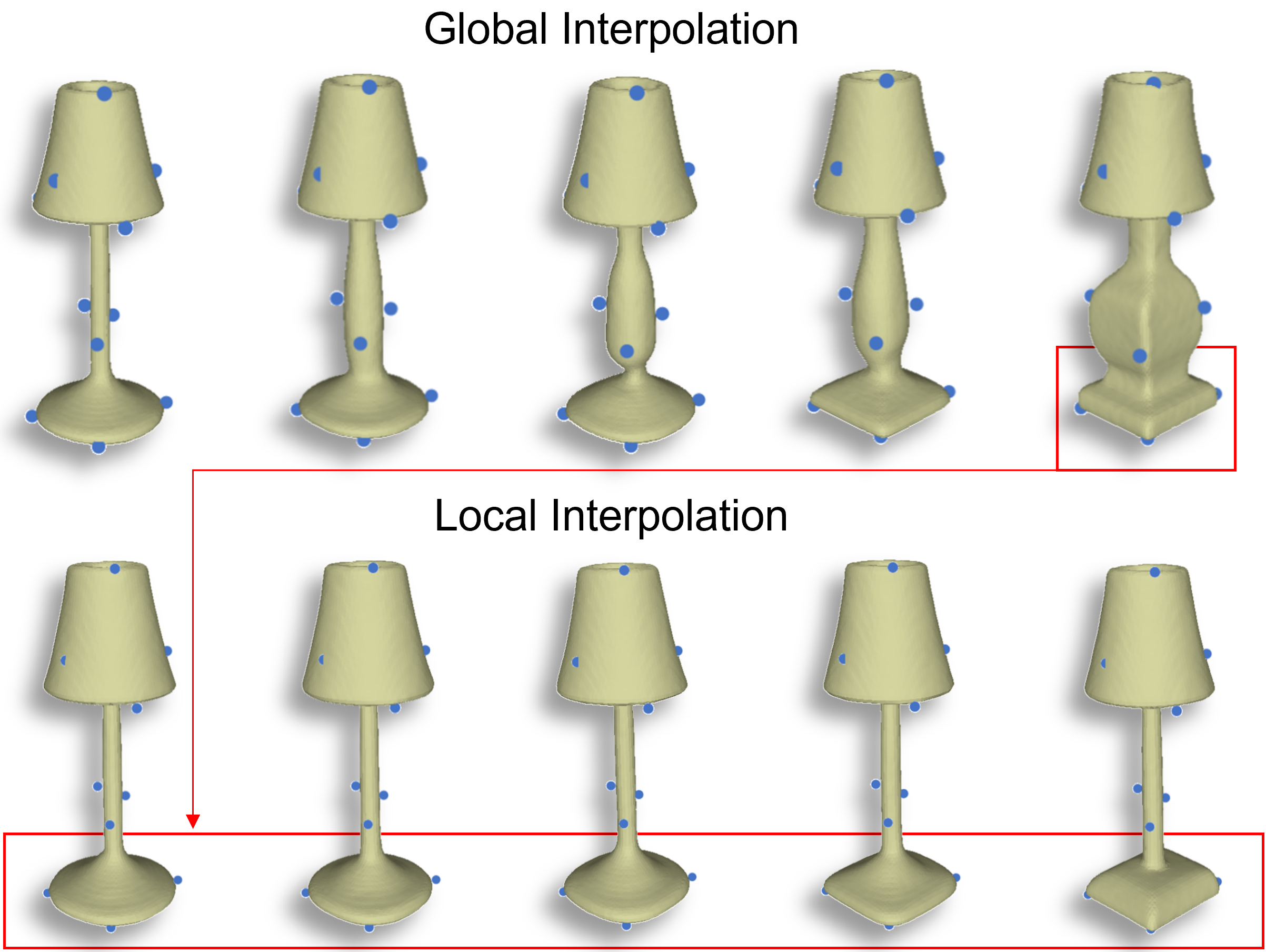}
    \vspace{-0.5em}
    \caption{SLIDE is able to perform both global and local interpolations. The first row is an example of global interpolation. The second row interpolates between the bottom of the two lamps.}
    \label{fig:interpolation}
\vspace{-1em}
\end{figure}
\vspace{-1.2em}
\paragraph{Shape Interpolation.}
To interpolate two shapes,
we can interpolate both the positions and features between the corresponding latent points of the two shapes.
See Appendix B.11 for how to establish correspondence between two sets of sparse latent points of two shapes.
The top row of Figure~\ref{fig:interpolation} is an example of global interpolation.
SLIDE is also able to perform local interpolation.
We can interpolate only a part of the latent points, and keep the positions and features of the rest part of the latent points fixed.
The bottom row of Figure~\ref{fig:interpolation} is an example of local interpolation.
\vspace{-1.2em}
\paragraph{Shape combination.}
We can also perform shape combinations using our sparse latent point-based representation of 3D shapes.
We can simply combine the sparse latent points and their features from two or more source shapes to form new shapes.
See Figure~\ref{fig:shape_combination} for an example.



\section{Conclusion}
\label{sec:conclusion}
In this work, we propose to use point clouds as an intermediate representation of meshes.
We train generative models on the point clouds sampled from the surface of the meshes, then we use SAP to reconstruct meshes from the generated point clouds.
Meshes generated in this way demonstrate diverse topology.
We propose to further encode dense point clouds to features at a sparse set of latent points, and train two DDPM in this latent space to learn the distribution of the positions and features of the latent points, respectively.
Our sparse latent point diffusion model (SLIDE) outperforms DDPMs directly trained on point clouds in terms of both sample quality and generation speed.
In addition, this sparse latent point representation allows us to explicitly control the shape of generated shapes, perform both global and local interpolations, and shape combination.


{\small
\bibliographystyle{ieee_fullname}
\bibliography{egbib}
}

\end{document}